%% file: root.tex
\documentclass[letterpaper, 10 pt, conference]{ieeeconf}  % Comment this line out if you need a4paper

\IEEEoverridecommandlockouts                              % This command is only needed if 
\usepackage{graphics} % for pdf, bitmapped graphics files
\usepackage{mathptmx} % assumes new font selection scheme installed
\usepackage{times} % assumes new font selection scheme installed
\usepackage{amsmath} % assumes amsmath package installed
\usepackage{amssymb}  % assumes amsmath package installed

\usepackage[nolist,nohyperlinks]{acronym}
\usepackage[utf8]{inputenc}
\usepackage{bm}
\usepackage{color}
\usepackage{graphicx}
\usepackage[noend]{algorithmic}
\usepackage{algorithm}
\let\labelindent\relax % necessary to avoid error from enumitem related to \labelindent being defined
\usepackage{enumitem}
\PassOptionsToPackage{hyphens}{url}\usepackage[hidelinks]{hyperref}
\usepackage{tablefootnote}
\usepackage{fontawesome5}
\usepackage{tikz}
\usetikzlibrary{shapes.geometric, arrows, shapes.multipart}

\title{\LARGE \bf
An Industrial Perspective on Multi-Agent Decision Making for Interoperable Robot Navigation following the VDA5050 Standard
}

\author{Niels van Duijkeren, Luigi Palmieri, Ralph Lange and Alexander Kleiner% <-this % stops a space
\thanks{All authors are with Robert Bosch GmbH, Corporate Research, Stuttgart, Germany%
        {\tt\small \{niels.vanduijkeren, luigi.palmieri, ralph.lange, alexander.kleiner\}@bosch.com}
        }%
}

\usepackage{color}

\newtheorem{remark}{Remark}

\newtheorem{researchquestion}{Research Opportunity}

\acrodef{AGV}[AGV]{Automated Guided Vehicle}
\acrodefplural{AGV}[AGVs]{Automated Guided Vehicles}
\acrodef{AMR}[AMR]{Autonomous Mobile Robot}
\acrodefplural{AMR}[AMRs]{Autonomous Mobile Robots}
\acrodef{MPC}[MPC]{Model Predictive Control}
\acrodef{PDDL}[PDDL]{Planning Domain Definition Language}
\acrodef{ASP}[ASP]{Answer Set Programming}
\acrodef{STRIPS}[STRIPS]{Stanford Research Institute Problem Solver}
\acrodef{MAP}[MAP]{Multi-Agent Planning}
\acrodef{MAPF}[MAPF]{Multi-Agent Path Finding}
\acrodef{MAPD}[MAPD]{Multi-Agent Pickup and Delivery}
\acrodef{AD}[AD]{Action Dependency}
\acrodef{ADG}[ADG]{Action Dependency Graph}
\acrodef{SADG}[SADG]{Switchable Action Dependency Graph}
\acrodef{BT}[BT]{Behavior Tree}
\acrodef{MILP}[MILP]{Mixed-Integer Linear Programming}
\acrodef{NLP}[NLP]{Non-Linear Programming}
\acrodef{LTL}[LTL]{Linear Temporal Logic}
\acrodef{OR}[OR]{Operations Research}
\acrodef{IoT}[IoT]{Internet of Things}
\acrodef{PLC}[PLC]{Programmable Logic Controller}
\acrodef{MQTT}[MQTT]{Message Queuing Telemetry Transport}
\acrodef{NURBS}[NURBS]{Non-Uniform Rational Basis Spline}
\acrodef{CBS}[CBS]{Conflict-Based Search}
\acrodef{SIPP}[SIPP]{Safe Interval Path Planning}
\acrodef{MAS}[MAS]{Multi-Agent System}
\acrodefplural{MAS}[MASs]{Multi-Agent Systems}

\newcommand\passref[1]{#1}
\tikzstyle{hzspec} = [font=\footnotesize, text=black!70]
\tikzstyle{coordstack} = [rectangle split, 
                          rectangle split parts=5,
                          rectangle split draw splits=false,
                          rectangle split part fill={gray!30,white},
                          rounded corners,
                          font=\footnotesize,
                          text centered,
                          draw]
\tikzstyle{navstack} = [rectangle split,
                        rectangle split parts=6,
                        rectangle split draw splits=false,
                        rectangle split part fill={gray!30,white},
                        rounded corners,
                        font=\footnotesize,
                        text centered,
                        draw]
\tikzstyle{arrow} = [thick,<->,>=stealth,dashed,thin]

\begin{document}

\maketitle
\thispagestyle{empty}
\pagestyle{empty}

%%%%%%%%%%%%%%%%%%%%%%%%%%%%%%%%%%%%%%%%%%%%%%%%%%%%%%%%%%%%%%%%%%%%%%%%%%%%%%%%
\begin{abstract}
  This paper provides a perspective on the literature and current challenges
    in \aclp{MAS} for interoperable robot navigation in industry.
  The focus is on the multi-agent decision stack for \aclp{AMR} operating
    in mixed environments with humans, manually driven vehicles, and legacy \aclp{AGV}.
  We provide typical characteristics of such \aclp{MAS} observed today and how these
    are expected to change on the short term due to the new standard VDA5050
    and the interoperability framework OpenRMF.
  We present recent changes in fleet management standards and the role of
    open middleware frameworks like ROS2 reaching industrial-grade quality.
  Approaches to increase the robustness and performance of multi-robot navigation
    systems for transportation are discussed, and research opportunities are derived.
\end{abstract}

%%%%%%%%%%%%%%%%%%%%%%%%%%%%%%%%%%%%%%%%%%%%%%%%%%%%%%%%%%%%%%%%%%%%%%%%%%%%%%%%
\input{contents/00_introduction.tex}
\input{contents/01_industrial_mas.tex}
\input{contents/02_single_agent_stack.tex}
\input{contents/03_multi_agent_stack.tex}
\input{contents/04_integration_and_benchmarking.tex}
\input{contents/98_conclusions.tex}
\input{contents/99_appendix.tex}

\addtolength{\textheight}{-0.0cm}   % This command serves to balance the column lengths

%%%%%%%%%%%%%%%%%%%%%%%%%%%%%%%%%%%%%%%%%%%%%%%%%%%%%%%%%%%%%%%%%%%%%%%%%%%%%%%%

%%%%%%%%%%%%%%%%%%%%%%%%%%%%%%%%%%%%%%%%%%%%%%%%%%%%%%%%%%%%%%%%%%%%%%%%%%%%%%%%

%%%%%%%%%%%%%%%%%%%%%%%%%%%%%%%%%%%%%%%%%%%%%%%%%%%%%%%%%%%%%%%%%%%%%%%%%%%%%%%%
% \section*{ACKNOWLEDGMENT}

%%%%%%%%%%%%%%%%%%%%%%%%%%%%%%%%%%%%%%%%%%%%%%%%%%%%%%%%%%%%%%%%%%%%%%%%%%%%%%%%

\bibliographystyle{ieeetr}
\bibliography{bib/literature}

\end{document}

%% file: contents/00_introduction.tex
%!TEX root = ../root.tex
\section{INTRODUCTION}\label{sec:intro}

\begin{figure}
  \centering
  \includegraphics[width=\columnwidth]{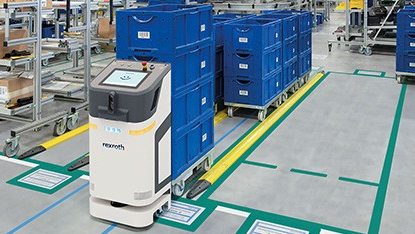}
  \caption{The ActiveShuttle by Bosch Rexroth, an \ac{AMR} for the automotive supply chain.}
  \label{fig:activeshuttle}
\end{figure}

Multi-robot navigation encompasses an ever-tighter integration of a vast number of disciplines
  and research as in most of the robotics research.
It is largely about the decision-making stack from translating high-level tasks into individual actuator
  commands for multiple robots \cite{Belta2007}.
Where \acp{AGV} have been the main go-to solution for a very long time,
  \acp{AMR} are rapidly gaining popularity driven by their flexibility
  and increasingly advanced capabilities.
An example of such an \ac{AMR} is the ActiveShuttle by Bosch Rexroth depicted in Fig.~\ref{fig:activeshuttle},
  primarily deployed for the automotive industry supply chain.
Another example are autonomous cleaning machines for e.g., offices and supermarkets that are currently coming onto the market.
Although these use cases are completely different,
  the decision-making stacks have much in common.

In this paper we present an industrial perspective on research on decision making
  in multi-agent systems and the current challenges for a settings in \emph{mixed environments},
  namely transportation tasks in factories.
Mixed environments here refer to the setting where the environment is shared with other
  ``uncontrolled'' agents such as humans and legacy (disconnected) \acp{AGV}.
The requirements for a multi-agent systems are, as usual, motivated in a top-down fashion.
We will however present the current view on the decision-making stack in a bottom-up order,
  that is from a single-agent behavior and motion planning stack to the task assignment problem.

Transportation tasks in factories are commonly to provision production machinery and assembly lines,
  either from storage or directly between different machines,
  and to take finalized components to storage locations.
The factory is structured by driving lanes that, as mentioned, are shared between vehicles and humans.

In a multi-agent setting,
  one robot is typically assigned to one transportation task at a time and the robots share
  the available resources (primarily space, but also infrastructure like charging stations).
The requirements from an operator perspective for industrial transportation systems is
  that they are efficient and reliable.
Higher efficiency helps keeping costs low by requiring fewer robots.
Reliability is needed to be able to effectively plan tasks for the \ac{MAS},
  and thereby the overall factory operations.
Reliability implies robustness against expected disturbances and uncertainties,
  however can practically also refer to the ability to provide timely status updates and thereby facilitate
  reactive planning.
Notice that these top-down requirements propagate through the entire decision-making stack.

Industry is steadily evolving towards more integrated planning and control mechanisms
  with the advent of Industry 4.0.
Production machinery in factories are adopting \ac{IoT} principles such as open APIs to facilitate this.
\acp{AGV} have been abundant for several decades,
  initially blindly following guidance systems in the floor.
The next trend were stand-alone \acp{AMR} with an isolated/proprietary form of coordination,
  but more flexible due to e.g., laser-based navigation. 

Now new vendor-independent protocols and standards for communication between fleet management and single robots
  such as VDA5050 \cite{VDA5050-2-0} and OpenRMF \cite{OpenRMF} are being developed.
VDA5050 is a standard pushed by the German Automotive-Industry Association (VDA) and the 
  German Mechanical Engineering Industry Association (VDMA) and is breaking open the market for
  fleet management and control of \acp{AMR} and creating new business opportunities.
Similarly, with ROS2 an open industrial-grade framework for building robotic systems is available,
  and by the Nav2 stack \cite{Macenski2020} and OpenRMF \cite{OpenRMF} also single-robot navigation
  and fleet management tools are provided.

\paragraph*{Contributions}
\emph{(a)} We present typical qualitative problem dimensions faced in industrial \acp{MAS}.
\emph{(b)} A reference industrial multi-agent decision-making stack is presented with currently common solution approaches.
Furthermore, we survey how recent developments in literature fit in and from that derive \emph{research opportunities}.
\emph{(c)} The role of the VDA5050 standard for industrial \acp{MAS} is discussed and so are the opportunities
  arising from ROS2 and the accomanying tools.

\paragraph*{Paper Structure}
In Section~\ref{sec:industrial_mas} we present a reference transportation case used in this paper.
Subsequently, in Section~\ref{sec:single_agent_stack} the single-agent decision stack is briefly
  discussed focussing on mechanisms and challenges to achieve robustness.
Thereafter, in Section~\ref{sec:multi_agent_stack} the centralized multi-agent coordination pipeline is
  detailed.
In Section~\ref{sec:integration_and_benchmarking} we discuss integration aspects and developments therein,
  and the role of benchmarking such integrated systems.
We conclude the paper with a short summary and research opportunities in Section~\ref{sec:conclusions}.

%% file: contents/01_industrial_mas.tex
%!TEX root = ../root.tex
\section{INDUSTRIAL MULTI-AGENT SYSTEMS}\label{sec:industrial_mas}

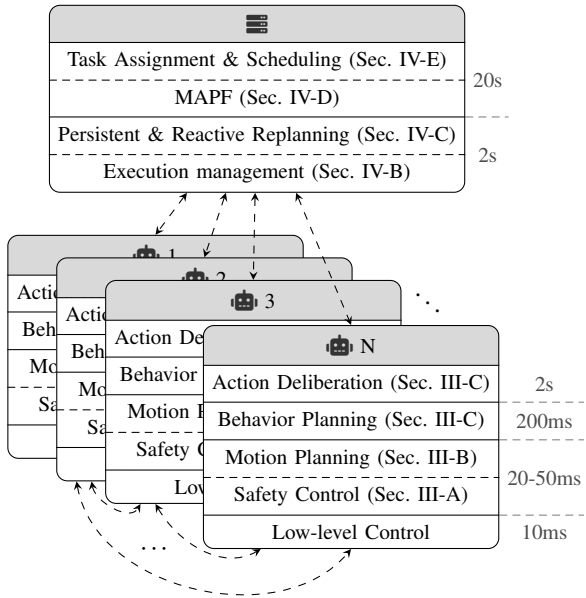
\begin{figure}
    \centering
    \input{figures/nav_coord_stack.tex}
    \caption{Schematic overview of a generic industrial multi-agent navigation stack with typical cycle times.}
    \label{fig:nav_coord_stack}
\end{figure}

Fig.~\ref{fig:nav_coord_stack} depicts a typical decision-making stack for autonomous navigation of a multi-agent
  robotic system in industry.
\begin{remark}
  We consider the same structure of the stack to be applicable for transportation and coverage tasks.
  The environments for coverage tasks are more diverse and still mostly an unexplored domain.
\end{remark}

We consider \acp{AMR} (possibly in groups of different types and vendors) operating
  in a semi-structured environment with a known roadmap of driving lanes.
The above system is effectively modeled by agents moving on a directed graph.
The agents operate in mixed environments,
  in which also other autonomous agents participate with uncertain behavior:

\vspace{3pt}

\paragraph*{Human-driven vehicles}
For example forklifts, pallet movers, and tugger trains.
\begin{itemize}
    \item The motions of these vehicles are constrained and assumed predictable.
    \item The vehicles are often significantly faster than \ac{AMR}s,
            it is therefore desirable that \ac{AMR}s let human-driven vehicles pass if the situation allows.
    \item The vehicles may occasionally halt on the driving lanes to load and unload.
            Although undesirable, the vehicles may also be parked for extended periods of time and partially block lanes.
    \item Personnel is instructed to give priority to \ac{AGV}s at intersections and not to disturb their motion. 
\end{itemize}

\paragraph*{Human workers} 
Moving solo or in groups.
\begin{itemize}
    \item Behavior and motion is hard to predict reliably, 
            robots ought to move carefully around humans to minimize risk of causing injuries.
    \item Personnel is instructed to give way to \ac{AGV}s at all times.
\end{itemize}

\paragraph*{Legacy \ac{AGV}s and \ac{AMR}s} 
Groups of \ac{AGV}s and \ac{AMR}s that use a legacy non-cooperative fleet management system.
\begin{itemize}
    \item The systems are usually the least flexible,
            and it is desirable to not interfere with their motions.
    \item The systems are safe and will stop the for anything that blocks their path,
            \acp{AMR} might---unlike \acp{AGV}---bypass the obstacle.
\end{itemize}

\begin{table}
    \centering
    \caption{Currently typical dimensions for a multi-agent system for robot navigation in factories.}
    \label{tab:problem-dimensions}
    \begin{tabular}{c|c}
      Number of agents in plant & 50-200\\
      Number of agent types \tablefootnote{For transporting loads of different types and sizes;
                          vehicles are often from different vendors.} & 4-8\\
      AMR speed (m/s) & 0.7-1.5\\
      Duration of tasks (minutes) & 3-20\\
      Typical delay per task (minutes) & 0-5\\
      Time between tasks (minutes)
        \tablefootnote{Machine-provisioning tasks are usually planned one by one as they arrive,
                    if pre-planning of factory operations is possible it is common that tasks arrive in batches.} & 1+\\
    \end{tabular}
\end{table}

There is much potential in restructuring a warehouse or factory such that the environment is more certain \cite{Wurman2008},
  this also allows much faster vehicles if the environment is free from humans.
For many small to medium enterprises and existing factories,
  the trend is that automation is introduced more gradually and dealing with uncertainty is highly relevant.
Fortunately the problem setting reveals enough structure such that a decomposition in subtasks is often trivial,
  and with easily-verifiable assumptions guarantees can be obtained.
Moreover,
  as Table~\ref{tab:problem-dimensions} reveals,
  the typically observed problem dimensions are manageable.

%% file: figures/nav_coord_stack.tex
\begin{tikzpicture}
    %% EDGE SERVER
    \node (coord) [coordstack] {
        \nodepart{one}\textcolor{black!80}{\faServer}
        \nodepart{two}Task Assignment \& Scheduling \passref{(Sec. \ref{sec:task_alloc})}
        \nodepart{three}MAPF \passref{(Sec. \ref{sec:mapf})}
        \nodepart{four}Persistent \& Reactive Replanning \passref{(Sec. \ref{sec:reactive_replanning})}
        \nodepart{five}Execution management \passref{(Sec. \ref{sec:execution_management})}
    };
    \draw (coord.one split west) -- (coord.one split east);
    \draw[densely dashed] (coord.two split west) -- (coord.two split east);
    \draw (coord.three split west) -- (coord.three split east);
    \draw[densely dashed] (coord.four split west) -- (coord.four split east);
    
    %% HZ FOR SERVER
    \def\ServerHzText{0.3cm}
    \def\ServerHzLength{0.7cm}
    \node (coord_two_east) [xshift=\ServerHzLength] at (coord.two split east) {};
    \node (coord_three_east) [xshift=\ServerHzLength] at (coord.three split east) {};
    \draw [gray,densely dashed] (coord.three split east) -- (coord_three_east);
    \node (mapf_rate) [hzspec, xshift=\ServerHzText, align=left] at (coord.two split east) {20s};
    \node (coord_rate) [hzspec, xshift=\ServerHzText, align=left] at (coord.four split east) {2s};

    %% ROBOTS
    \def\SkipRobot{4}
    \def\NumRobots{5}
    \node (robot0) [below of=coord, xshift=-2.0cm, yshift=-2.0cm]{};
    \foreach \x [count=\i from 0] in {1,...,\NumRobots} {
        \ifthenelse{\x = \SkipRobot}{
            \node (robot\x) [below of=robot\i, xshift=0.65cm, yshift=0.7cm] {};
            \node (robotN) [xshift=0.35cm, yshift=-0.2cm] at (robot\i.north east) {$\ddots$};
        }{
            \ifthenelse{\x = \NumRobots}{
                \def\RobotName{N }
            }{
                \def\RobotName{\x}
            }
            \node (robot\x) [navstack, below of=robot\i, xshift=0.65cm, yshift=0.7cm] {
                \nodepart{one}\textcolor{black!80}{\faRobot{}} \RobotName
                \nodepart{two}Action Deliberation \passref{(Sec.~\ref{sec:behavior_planning_and_delib})}
                \nodepart{three}Behavior Planning \passref{(Sec.~\ref{sec:behavior_planning_and_delib})}
                \nodepart{four}Motion Planning \passref{(Sec.~\ref{sec:motion_planning})}
                \nodepart{five}Safety Control \passref{(Sec.~\ref{sec:safety_control})}
                \nodepart{six}Low-level Control
            };
            \draw (robot\x.one split west) -- (robot\x.one split east);
            \draw (robot\x.two split west) -- (robot\x.two split east);
            \draw (robot\x.three split west) -- (robot\x.three split east);
            \draw[densely dashed] (robot\x.four split west) -- (robot\x.four split east);
            \draw (robot\x.five split west) -- (robot\x.five split east);
            \draw [arrow] (coord)--(robot\x.north);
        }
    }

    %% ARROWS BETWEEN ROBOTS
    \draw [arrow] ([xshift=-1.3cm]robot3.south) to [out=300,in=210] ([xshift=-1.2cm]robot5.south);
    \draw [arrow] ([xshift=-1.7cm]robot2.south) to [out=280,in=230] (robot5.south);
    \draw [arrow] ([xshift=-1.5cm]robot2.south) to [out=280,in=240] ([xshift=-1.5cm]robot3.south);
    \node (arrows) [xshift=0.0cm, yshift=-1.2cm] at (robot1.south) {$\cdots$};

    %% HZ FOR ROBOT
    \def\RobotHzText{0.6cm}
    \def\RobotHzLength{1.25cm}
    \node (robot_two_east) [xshift=\RobotHzLength] at (robot\NumRobots.two split east) {};
    \node (robot_three_east) [xshift=\RobotHzLength] at (robot\NumRobots.three split east) {};
    \node (robot_four_east) [xshift=\RobotHzLength] at (robot\NumRobots.four split east) {};
    \node (robot_five_east) [xshift=\RobotHzLength] at (robot\NumRobots.five split east) {};
    \draw [gray,densely dashed] (robot\NumRobots.two split east) -- (robot_two_east);
    \draw [gray,densely dashed] (robot\NumRobots.three split east) -- (robot_three_east);
    \draw [gray,densely dashed] (robot\NumRobots.five split east) -- (robot_five_east);
    \node (strategy_planning_rate) [hzspec, xshift=\RobotHzText, align=left] at (robot\NumRobots.two east) {2s};
    \node (behavior_rate) [hzspec, xshift=\RobotHzText, align=left] at (robot\NumRobots.three east) {200ms};
    \node (motion_rate) [hzspec, xshift=\RobotHzText, align=left] at (robot\NumRobots.four split east) {20-50ms};
    \node (lowlevel_rate) [hzspec, xshift=\RobotHzText, align=left] at (robot\NumRobots.six east) {10ms};
\end{tikzpicture}

%% file: contents/02_single_agent_stack.tex
%!TEX root = ../root.tex
\section{SINGLE-AGENT PLANNING STACK}\label{sec:single_agent_stack}

The goal for the single-agent planning stack is to execute short motion tasks with a
  horizon in the order of up to 10 seconds.
A single-agent planning stack provides information about the robot capabilities and
  constraints.
The stack needs to ensure that a task is executed safely and reliably and needs to be able to provide feedback
  in case task execution is delayed or stuck and provide possible solutions.

\subsection{Safety Control}\label{sec:safety_control}

In mixed environments,
  the layer that connects motion planning with the low-level actuators is a safety control unit.
Often a dedicated \ac{PLC} is paired with one or multiple certified safety Lidar scanners
  to monitor that pre-programmed, velocity-dependent, areas remain clear of obstacles and the \ac{PLC} intervenes if they do not.
More modern approaches aim to not employ discrete switching behavior anymore \cite{Wabersich2021}.
\acp{AGV} rely on these systems to brake for obstacles in the way,
  the aim for \acp{AMR} is to keep these areas free of obstacles in the motion planning stage.
Safety controllers override any motion commands if necessary and safety constraints have a significant influence
  on the robot behaviors and efficiency.

\subsection{Motion Planning}\label{sec:motion_planning}

The task of the motion planner is to drive the robot fast and smooth to the given goal.
This layer concerns itself with the vehicle dynamics, action limits,
  load stability, and user-defined bounds on velocity and acceleration.
In principle a reference path can be assumed to be given,
  however an explicitly-specified maximum deviation serves to allow avoiding obstacles by planning in a corridor \cite{VanDuijkeren2019}.
\ac{MPC} is a popular approach to the motion planning problem with enabling tools freely available \cite{Verschueren2021}.
Through extended models it is avoided that robots carrying heavy loads require excessive currents to drive due to the orientation of caster wheels \cite{Arrizabalaga2021}.
Motion planning is typically restricted to making decisions in continuous spaces with ideally differentiable
  objectives and constraints.

\subsection{Behavior Planning and Action Deliberation}\label{sec:behavior_planning_and_delib}

Behavior planning in contrast is concerned with making logical decisions,
  such as whether to yield at an intersection, overtake vehicles, or facilitate being overtaken.
In the architecture of Fig.~\ref{fig:nav_coord_stack} the function
  is separated in two layers:
  (1) a behavior control layer that implements reactive behavior and executes sequences of actions,
  (2) an action deliberation layer that considers the interaction with other agents,
      the interaction with the multi-agent planning stack, and plans action sequences to optimize performance.
For behavior planning,
  behavior trees have rapidly spread in from the gaming industry to robotics \cite{Iovino2022}.
They are a flexible representation of robot behaviors that can be programmed by hand,
  synthesized from formal methods such as \ac{LTL} \cite{Colledanchise2017},
  or obtained from reinforcement learning or learning from demonstration.
The action deliberation layer can for instance be the
  high-level scoring and validation of several possible scenarios,
  committing to the best scenario and steer the behavior planning accordingly.

Although we will not enter into much detail of this aspect,
  but we note that humans might interfere with the tasks that were allocated by the multi-agent planning stack.
Robots typically have multiple modes of user interaction that override the autonomous task execution.

\subsection{Distributed Coordination and Planning}\label{sec:distr_coordination}

There exist many works on distributed coordination and cooperative planning for most
  of the above functional layers and may help to increase robustness in multi-agent plan execution.
Approaches include precisely coordinating continuous trajectories \cite{VanParys2017}
  and avoid collisions between agents while deviating from the prescribed path \cite{Firoozi2020}.
Distributed negotiation over resources becomes relevant when simultaneous movements are not
  possible due to resource constraints \cite{Liu2018}.

This type of coordination is sometimes referred to as \emph{swarm intelligence} \cite{Agilox2021},
  but unfortunately no open standards have been developed for this purpose yet.
This functionality allow robots to for instance collaboratively execute tasks that could
  not be performed by a single robot.

%% file: contents/03_multi_agent_stack.tex
%!TEX root = ../root.tex
\section{MULTI-AGENT PLANNING STACK}\label{sec:multi_agent_stack}

The multi-agent planning stack is responsible for orchestrating the motions of single agents
  such that high-level tasks are completed.
In this paper one can assume that the transportation or cleaning tasks are already decomposed such
  that they can principally be executed by single agents.

\subsection{VDA5050 Standard for Transportation}

The VDA5050 standard for fleet management is developed and promoted by major industry associations in Germany.
Its primary scope is to control wheeled \acp{AMR} moving in a 2D model of the world.
The text of the standard is available on a GitHub repository and contributions can be made via pull requests \cite{VDA5050-2-0}.
The VDA5050 protocol assumes there is one of a so-called \emph{master control} that controls all vehicles in
  a plant.
The standard is primarily a definition of the communication between the master control and the
  \acp{AGV}/\acp{AMR}.
Hereby, the vehicles of different vendors can be planned and controlled by a single fleet management system.
Similarly, the fleet management system can be freely replaced by another if desired.
There are facilities in the standard to allow intermediate proprietary control layers for legacy systems,
  for this it is assumed that all relevant information of the vehicles is relayed to the VDA5050-based control.
The aim of the standard is to support \acp{MAS} consisting of up to  a few thousand vehicles.

The VDA5050 standard prescribes usage of \ac{MQTT} protocol as a means of communication
  via wireless networks and the messages are packed in a JSON format.
The plant operator defines a roadmap with constraints for the edges and properties of the nodes.
Each \ac{AGV} communicates a factsheet about its properties and abilities.
The \acp{AGV} are expected to be able to perform single-agent tasks and that they transmit their status
  frequently.
The \emph{master control} is responsible for \ac{MAP}, conflict resolution and traffic control,
  charging jobs, communication with the infrastructure (e.g., doors), and resolving communication problems. 

The control concept of VDA5050 is to send task orders to the agents.
These task orders are decomposed in two parts,
  the \emph{base} and the \emph{horizon}.
The base is the part of the task that the \ac{AGV} is expected to execute immediately and is not assumed to be cancellable.
The horizon is not yet confirmed and therefore only informative and may change.
A typical implementation will order a base that is quite short where the \ac{AGV} has to drive up to e.g. 6 meters,
  where the horizon may be the entire route to the goal or just a few edges beyond the base.
The order message consists of over 50 fields and includes the target position and orientation,
  the nodes to traverse, the actions to perform on each node, the shape of path between nodes as straight lines
  or by \ac{NURBS} curves, the maximum deviation (e.g., to allow driving around obstacles), the maximum velocity, 
  maximum height (for high loads), et cetera.
There is a predefined set of actions (pick, place, start/stop charging, etc.),
  but it is foreseen that vendors can extend these based on the needs of an application.
The status message sent by the robot includes the current state, but also informs the multi-agent planning stack
  and indirectly the operator about contingencies to e.g. diagnose delayed or stuck robots.

\subsection{Execution Management}\label{sec:execution_management}

Note that the VDA5050 is tailored to a specific, but abundant, 
  use case often seen in the automotive and mechanical engineering industry.
The standard focusses on the interaction between a central fleet manager and the robotic fleet,
  to define the means for orchestrating the robot motion,
  which naturally implies that a \ac{MAPF} component is present.
However, at the execution-management level of the planning and coordination stack,
  it is assumed that the desired trajectories for all robots are available.
The main responsibility is to ensure robust execution of these plans through
  real-time monitoring and allocation of physical resources to agents.
The procedure is typically implemented by a reactive resource-reservation system.
Based on the current state of the robot or fleet,
  one robot receives the clearance to move along a certain edge.
This may be as simple as a first-come-first-serve assignment,
  but more suitably a method that avoids deadlocks among the agents participating in the \ac{MAPF} plan \cite{Ma2017a,Hoenig2019,Atzmon2020}.
Inspired by optimization-based street traffic management approaches \cite{Lin2011},
  one might consider dealing with deadlocks and congestions implicitly but thereby account for humans \cite{Rudenko2020} and autonomous agents too.
On a lower level,
  execution management optimizations are on precisely timing how agents pass through intersections \cite{Dresner2008,Hult2015}.
Finally,
  in particular when robots need to temporarily deviate from their assigned corridors,
  for instance to circumvent blockages or to overtake slower vehicles,
  coordination on the level of continuous trajectories is appropriate \cite{Pecora2018}.

\subsection{Persistent \& Reactive Replanning}\label{sec:reactive_replanning}

Unmodeled disturbances in the execution of an \ac{MAPF} plan can have mild consequences on the efficiency of the
  multi-agent system but may also be dramatic in case a deadlock occurs.
Motivated hereby,
  the \ac{ADG} was proposed to maintain and enforce an execution schedule
  that can be derived from most \ac{MAPF} plans \cite{Hoenig2019}.
A follow-up work presents the so-called \ac{SADG} and a method to optimally change the order of planned actions 
  in the \ac{MAPF} to react on disturbances while guaranteeing to preserve deadlock-freeness \cite{Berndt2020}.
There is even more potential in promptly rerouting vehicles in the event of unforeseen congestions \cite{Vedder2021}.
One can consider the problem as to continuously locally repairing \ac{MAPF} plans \cite{Felner2007},
  but also to continuously process incoming jobs such as in \cite{Ma2017}.
All the above is ideally executed at the time scales of several seconds,
  a rapid response is critical and temporary suboptimality is rather unproblematic.
\ac{CBS} methods \cite{Sharon2015} seem naturally suited for continuous updates
  reactively replanning based on observed real conflicts as suggested in e.g. \cite{Atzmon2020}.

\subsection{Multi-Agent Path Finding}\label{sec:mapf}

The goal of the \ac{MAPF} component is to find approximately optimal routes for tasks that are not yet being executed.
These tasks are typically in a queue of tasks to be started from about half a minute up to 30 minutes in the future.
The \ac{MAPF} is presented separately from task assignment because it remains to be commonly implemented this way.
\ac{MAPF} is a well-established research area \cite{Stern2019} and has a plethora of highly performant algorithms.
The algorithms are sufficiently fast for problem sizes presented in Tab.~\ref{tab:problem-dimensions} and for the 
  scaling expected over the next few years and the scope of VDA5050.
Optimality in \ac{MAPF} is usually defined as either the plans that minimize the \emph{makespan} or those that
  minimizes the \emph{sum-of-costs}.
The latter is often chosen for the heuristic that it is typically appreciated that any single robot finishes
  their task sooner if possible to be available for new tasks.
The \ac{MAPF} problem is often still approached by the classical prioritized planning methods \cite{Erdmann1987}
  and single-agent plans are generated sequentially in the order that the paths for tasks are being planned.
For larger scale use cases practically improved completeness \cite{Cap2015},
  or complete methods such as \ac{SIPP} \cite{Phillips2011}, \ac{CBS} \cite{Sharon2015} are appreciated.
So is the speed-up that is attained through follow-up contributions \cite{Barer2014,Boyarski2015,Cohen2016a},
  and very recently \cite{Li2022}.
It is remarked that concepts where plant operators can improve the search by defining e.g.,
  highways \cite{Cohen2016a} can be very useful in practice if properly exploited.
Robust path-finding methods and the framework presented in \cite{Atzmon2020} are highly relevant to the industrial
  use case and form a basis for the work presented here.
Finally, we note that methods to combine data-driven models for heuristics with suboptimal search
  hold a lot of potential \cite{Spies2019}.

\subsection{Task Assignment and Scheduling}\label{sec:task_alloc}

Task assignment is the problem of deciding which agent should execute which task in the queue,
  or if necessary what combination of robots ought to be used.
One straightforward approach is to select the robot that can reach the starting position of the task the soonest.
This approach often works quite well,
  but suffers from the usual shortcomings of a greedy decision-making approach.
It does confirm that for transportation use cases the task assignment problem is 
  closely related to the \ac{MAPF} problem.
It is therefore no surprise that it is proposed to solve the optimal \ac{MAPF} and task-assignment
  problem simultaneously \cite{Hoenig2018}.
Task assignment and scheduling interface with cooperative multi-agent task planning mechanisms upwards \cite{Khamis2015,Torreno2018},
  these methods are out of scope for this paper but are becoming increasingly relevant for industry.

%% file: contents/04_integration_and_benchmarking.tex
%!TEX root = ../root.tex
\section{INTEGRATION AND BENCHMARKING}\label{sec:integration_and_benchmarking}

The integration of a complex robotic system is a challenging engineering endeavor.
Combining many sensors, actuators, communication hardware, and complex software is not a new domain.
However, these robots operate with little human supervision in uncertain environments,
  while they are expected to autonomously resolve conflicts in a rationally legible manner.
All required technologies to make this happen move forward with a fast pace,
  and testing and measuring these advances in an integrated system is essential.
It can be frustrating to realize that 
  one little thing that goes wrong in the decision-making stack,
  may quickly result in classical \acsp{AGV} to outperform the more modern systems.

\subsection{Robot Operating System 2 (ROS2)}

ROS and ROS2 are the de-facto standard frameworks in academia for developing any robotic system,
  with ROS2 it steadily gains traction in industry to fulfill that same role.
The core of ROS2 is to provide a framework for intra- and inter-robot communication and interaction.
It is complemented with a large ecosystem of algorithms and tools to quickly implement,
  simulate, and deploy robotic systems.
The active involvement of industry in the development of ROS2 has lead to a very high software-quality
  standard, it becomes increasingly rare that a problem in the ROS2 core is a cause for problems in a robotic system.
The same is true for the most popular tools in ROS2, such as
\emph{(a)} \emph{Navigation2}, the highly flexible plugin-based stack for single-robot navigation \cite{Macenski2020};
\emph{(b)} \emph{PlanSys2}, a PDDL-based planning system to conveniently write,
  deploy, and troubleshoot AI-planning based decision-making algorithms \cite{Martin2021};
\emph{(c)} \emph{OpenRMF}, a framework for implementing fleet management and multi-agent planning systems \cite{OpenRMF}.

OpenRMF is a relatively new project that is not based on an already popular ROS1 counterpart like Navigation2 and PlanSys2 had.
It brings new tools to build web interfaces, connecting to automated task dispatching systems,
  connecting to entire fleet management systems or single devices, and the core components of
  facilitating auctions, traffic coordination, and scheduling.
Any useful tool implemented in OpenRMF is expected to gain rapid adoption in both academia and perhaps eventually industry.

Another relevant ROS2 community effort highly relevant to this paper is the \texttt{vda5050\_msgs} package \cite{vda5050-msgs} 
  to conveniently couple the VDA5050 standard with ROS2 via an \ac{MQTT} bridge.

\subsection{Benchmarking}

Whether your robotic problem is really solved by an engineered solution is hard to decide.
Bringing a competitive product to the market ultimately means that the costs of the solution are justifiable for the customer.
The desirable high efficiency and reliability are two key criteria which one can attempt to measure.
However,
  increasing modularization and open standards like VDA5050 make this problem hard to do right.
In practice,
  a multi-agent planning stack should work well with any mixture of robot teams from ideally every possible combination of vendors.
The development of highly-automated simulation-based benchmarking suites with an ever-growing portfolio of cases and scenarios is indispensible.
In order to avoid that every company needs to repeat these efforts,
  especially open-source activities in this direction are highly valuable \cite{mrp-bench}.
This latter open-source project based on ROS2 and OpenRMF was recently started and hopefully gathers momentum in the near future.

%% file: contents/98_conclusions.tex
%!TEX root = ../root.tex
\section{CONCLUSIONS AND OPPORTUNITIES}\label{sec:conclusions}

In this paper we gave a perspective on \aclp{MAS} for industrial navigation.
By presenting the problem setting in detail we could confirm the industrial relevance
  of many recent advances in research.
We would like to conclude this paper with a list of application-driven research challenges we have identified:

\begin{researchquestion}
  The main goal for the persistent replanning function detailed in Sec.~\ref{sec:reactive_replanning} is to maintain an
    optimized flow on the roadmap graph.
  How can controlled agents and uncertain uncontrolled agents be optimally guided to avoid congestions and deadlocks?
  What role can data-driven models play to this end?
\end{researchquestion}
\begin{researchquestion}
  We present \ac{MAPF} (Sec.~\ref{sec:mapf}) to mainly be about planning new paths and replanning (Sec~\ref{sec:reactive_replanning}) 
    about dealing with disturbances.
  How can these steps be arranged such that they operate on a common plan and e.g.,
    existing plans are modified based on new task information.
\end{researchquestion}
\begin{researchquestion}
  Roadmap graphs considered in Sec.~\ref{sec:mapf} represent a restricted view on physically possible motions and in 
    case of a blockage, an \ac{AMR} may be able to drive around it.
  How can we model temporary free space motions in a persistent \ac{MAPF} planning scheme?
\end{researchquestion}
\begin{researchquestion}
  \ac{MAPF} planning in factories is highly repetitive,
    we can potentially obtain rich data-driven models to improve the quality of plans over time.
  How can such data-driven models be effectively used in the planning stage without imposing a restrictive computational burden. 
\end{researchquestion}
\begin{researchquestion}
  How can industrial companies cooperate on a common benchmarking platform to facilitate optimization of complementary products
    by other vendors without sharing sensitive intellectual property.
\end{researchquestion}

%% file: contents/99_appendix.tex
%!TEX root = ../root.tex